%% file: main.tex
\documentclass[11pt,a4paper]{article}
\usepackage{arabtex}
\usepackage{emnlp2018}
\usepackage{times}
\usepackage{latexsym}
\usepackage{amsmath}
\usepackage{amsfonts}
\usepackage{multirow}

\usepackage{url}
\usepackage{color}
\usepackage{graphicx}
\usepackage{textcomp}
\usepackage{fancyvrb}
\usepackage{umoline}
\usepackage{footnote}
\usepackage{tabularx, booktabs}
\usepackage{soul}
\usepackage{enumitem}
\usepackage{algpseudocode}
\usepackage{algcompatible}
\usepackage{subcaption}

\usepackage{linguex}
\usepackage{ulem}
\usepackage[utf8]{inputenc}
\usepackage{CJKutf8}
\usepackage[all]{nowidow}

\aclfinalcopy 

\newcolumntype{Y}{>{\centering\arraybackslash}X}

\newcommand{\tb}[1]{\textbf{#1}}
\newcommand{\tabitem}{~~\llap{\textbullet}~~}

\newcommand{\mname}{CMX}
\newcommand{\mfull}{\mname}
\newcommand{\msmall}{\mname-small}
\newcommand{\ourdata}{GY-Mix}
\newcommand{\kbdata}{KB-Mono56}
\newcommand{\webdata}{Web-Mix6}
\newcommand{\twitterdata}{Twitter-Mix}
\newcommand{\equilid}{EquiLID}
\newcommand{\lanidenn}{LanideNN}

\title{A Fast, Compact, Accurate Model for Language Identification of Codemixed Text}

\author{Yuan Zhang, Jason Riesa, Daniel Gillick, Anton Bakalov, Jason Baldridge, David Weiss \\
  Google AI Language \\
  {\tt \{zhangyua, riesa, dgillick, abakalov, jridge, djweiss\}@google.com} \\
}
\date{}

\begin{document}
\maketitle
\begin{abstract}
\input{abstract} 
\end{abstract}

\input{intro_jridge}
\input{data}

\input{model}

\input{results}
\input{related}
\input{conclusion}

\section*{Acknowledgments}
We thank Emily Pitler, Slav Petrov, John Alex, Daniel Andor, Kellie Webster, Vera Axelrod, Kuzman Ganchev, Jan Botha, and Manaal Faruqui
for helpful discussions during this work, and our anonymous reviewers for their
thoughtful comments and suggestions. We also thank Elixabete Gomez, H\'ector Alcalde, Knot Pipatsrisawat, and their stellar team of linguists who helped us to annotate and curate much of our data.

\bibliographystyle{acl_natbib}
\bibliography{paper}

\end{document}

%% file: abstract.tex

We address fine-grained multilingual language identification: providing a language code for every token in a sentence, including codemixed text containing multiple languages. Such text is prevalent online, in documents, social media, and message boards. We show that a feed-forward network with a simple globally constrained decoder can accurately and rapidly label both codemixed and monolingual text in 100 languages and 100 language pairs. This model outperforms previously published multilingual approaches in terms of both accuracy and speed, yielding an 800x speed-up and a 19.5\% averaged absolute gain on three codemixed datasets. It furthermore outperforms several benchmark systems on monolingual language identification.

%% file: intro_jridge.tex
\section{Introduction} \label{sec:intro}

Codemixed text is common in user-generated content, such as web articles, tweets, and message boards, but most current language ID models ignore it. Codemixing involves language switches within and across constituents, as seen in these English-Spanish and English-Hindi examples.

\ex. \label{ex:intramix} \dashuline{dame} [$_{NP}$ \dashuline{ese} \uline{book that you told me about}]\\
	{\it Give me this book that you told me about.}
    
\vspace{-0.2in}
\ex. \label{ex:intermix} [$_{NP}$ \dashuline{aapki} \uline{profile photo}] [$_{VP}$ \dashuline{pyari hai}]\\
     {\it Your profile photo is lovely.}


\noindent
Codemixing is the norm in many communities, e.g. speakers of both Hindi and English. As much as 17\% of Indian Facebook posts \citep{bali-EtAl:2014:CodeSwitch} and 3.5\% of all tweets \citep{rijhwani-EtAl:2017:Long} are codemixed. This paper addresses fine-grained (token-level) language ID, which is needed for many multilingual downstream tasks, including syntactic analysis \citep{Bhat:2018}, machine translation and dialog systems. Consider this example, which seeks a Spanish translation for the English word \textit{squirrel}:

\ex. \dashuline{como se llama un} \uline{squirrel} \dashuline{en español} \\
\textit{What do you call a squirrel in Spanish?}

Per-token language labels are needed; a system cannot handle the whole input while assuming it is entirely English or Spanish.

Fine-grained language ID presents new challenges beyond sentence- or document-level language ID. Document-level labels are often available in metadata, but token-level labels are not. Obtaining token-level labels for hundreds of languages is infeasible: candidate codemixed examples must be identified and multilingual speakers are required to annotate them. Furthermore, language ID models typically use character- and word-level statistics as signals, but shorter inputs have greater ambiguity and less context for predictions. Moreover, codemixing is common in informal contexts that often have non-standard words, misspellings, transliteration, and abbreviations \citep{Baldwin:2013}. Consider \ref{ex:fr-ar}, a French-Arabic utterance that has undergone transliteration, abbreviation and diacritic removal.

\novocalize
\ex. \label{ex:fr-ar} \dashuline{cv bien} \uline{hmd w enti}\\
     \dashuline{ça va bien} \uline{\textbf{al}h\textbf{a}md\textbf{ullilah} w\textbf{a} enti}\\
     \dashuline{ça va bien} \uline{\RL{al.hmd al-ll_ahi wAnt}}\\
 	{\it It's going well, thank God, and you?}     

\noindent
Language ID models must be fine-grained and robust to surface variations to handle such cases.


We introduce {\textbf \mname}, a fast, accurate language ID model for \tb{C}ode\tb{M}i\tb{X}ed text that tackles these challenges. \mname\ first outputs a language distribution for every token independently with efficient feed-forward classifiers. Then, a decoder chooses labels using both the token predictions and global constraints over the entire sentence. This decoder produces high-quality predictions on monolingual texts as well as codemixed inputs. We furthermore show how selective, grouped dropout enables a blend of character and word-level features in a single model without the latter overwhelming the former. This dropout method is especially important for \mname's robustness on informal texts.

We also create synthetic training data to compensate for the lack of token-level annotations. Based on linguistic patterns observed in real-world codemixed texts, we generate two million codemixed examples in 100 languages. In addition, we construct and evaluate on a new codemixed corpus of token-level language ID labels for 25k codemixed sentences (330k tokens). This corpus contains examples derived from user-generated posts that contain English mixed with Spanish, Hindi or Indonesian. 

Language ID of monolingual text has been extensively studied \citep{Hughes:2006,Baldwin:2010,Lui:2012,King:2013}, but language ID for codemixed text has received much less attention. Some prior work has focused on identifying larger language spans in longer documents \citep{Lui:2014,jurgens2017incorporating} or estimating proportions of multiple languages in a text \citep{Lui:2014,Kocmi:2017}. Others have focused on token-level language ID; some work is constrained to predicting word-level labels from a single language pair \citep{Nguyen:2013,solorio-EtAl:2014:CodeSwitch,molina-EtAl:2016:W16-58,Sristy:2017}, while others permit a handful of languages \citep{Das:2014,Sristy:2017,rijhwani-EtAl:2017:Long}. In contrast, \mname\ supports 100 languages. Unlike most previous work--with \citealt{rijhwani-EtAl:2017:Long} a notable exception--we do not assume a particular language pair at inference time. Instead, we only assume a large fixed set of language pairs as a general constraint for all inputs.

We define and evaluate \mname\ and show that it strongly outperforms state-of-the-art language ID models on three codemixed test sets covering ten languages, and a monolingual test set including 56 languages. It obtains a 19.5\% absolute gain on codemixed data and a 1.1\% absolute gain (24\% error reduction) on the monolingual corpus. Our analysis reveals that the gains are even more pronounced on shorter text, where the language ID task naturally becomes more difficult. In terms of runtime speed, \mname\ is roughly 800x faster than existing token-level models when tested on the same machine. Finally, we demonstrate a resource-constrained but competitive variant of \mname\ that reduces memory usage from 30M to 0.9M.


%% file: data.tex
\section{Data}
\label{sec:data}

We create synthetic codemixed training examples to address the expense and consequent paucity of token-level language ID labels. We also annotate real-world codemixed texts to measure performance of our models, understand code-mixing patterns and measure the impact of having such examples as training data. 

\paragraph{Synthetic data generation from monolingual text.}
For training models that support hundreds of languages, it is simply infeasible to obtain manual token-level annotations to cover every codemixing scenario \citep{rijhwani-EtAl:2017:Long}. However, it is often easy to obtain sentence-level language labels for monolingual texts.
This allows projection of sentence-level labels to each token, but a model trained only on such examples will lack codemixed contexts and thus rarely switch within a sentence. To address this, we create synthetic training examples that mix languages within the same sequence. 

To this end, we first collect a monolingual corpus of 100 languages from two public resources: the W2C corpus\footnote{\scriptsize \url{http://ufal.mff.cuni.cz/w2c}} and the Corpus Crawler project.\footnote{\scriptsize \url{https://github.com/googlei18n/corpuscrawler}} Then we generate a total of two million synthetic codemixed examples for all languages.

In generating each training example, we first sample a pair of languages uniformly.\footnote{Both our collected codemixed data and \citet{Barman:2014} indicate that more than 95\% of codemixed instances are bilingual.} We sample from a set of 100 language pairs, mainly  including the combination of English and a non-English language. The full set is listed in the supplemental material. Then we choose uniformly between generating an \textit{intra-mix} or \textit{inter-mix} example, which are two of the most prominent types of codemixing in the real world \cite{Barman:2014, Das:2014}.\footnote{The two types of codemixing have roughly equal proportions in our labeled corpus.} An \textit{intra-mix} sentence like \ref{ex:intramix} starts with one language and switches to another language, while an \textit{inter-mix} sentence like \ref{ex:intermix} has an overall single language with words from a second language in the middle. To generate an example, we uniformly draw phrases from our monolingual corpus for the chosen target languages, and then concatenate or mix phrases randomly. The shorter phrase in \textit{inter-mix} examples contains one or two tokens, and the maximum length of each example is eight tokens.

\paragraph{Manual annotations on real-world codemixed text.}
We obtain candidates by sampling codemixed public posts from  Google+\footnote{\scriptsize \url{https://plus.google.com/}} and video comments from YouTube,\footnote{\scriptsize \url{https://www.youtube.com/}} limited to three language pairs with frequent code switching: English-Spanish, English-Hindi\footnote{Hindi texts found in both Devanagari and Latin scripts.} and English-Indonesian. All texts are tokenized and lowercased by a simple rule-based model before annotation. Both the candidate selection and the annotation procedures are done by linguists proficient in both languages. The final annotated corpus contains 24.7k sentences with 334k tokens; 30\% are monolingual, 67\% are bilingual and 3\% have more than two languages. Finally, we create an 80/10/10 split (based on tokens) for training, development and testing, respectively. Table \ref{tb:statistics} gives the token and sentence counts per language. In the rest of the paper, we refer to this dataset as \ourdata.




\begin{table}[t]
    \centering
    \begin{tabular}{l|ccc}
     & en/es & en/hi & en/id\\
    \hline
    Number of tokens  & 98k & 140k & 94k \\
    Number of sentences  & 9.5k & 9.9k & 5.3k \\
    \end{tabular}
    \caption{Statistics of our YouTube and Google+ dataset, \ourdata.}
	\label{tb:statistics}
\end{table}

\begin{table}[t]
    \centering
    \begin{tabular}{r|l}
    Test Set & Languages \\
    \hline
    \twitterdata & en, es \\
    \webdata & cs, en, eu, hu, hr, sk \\
    \ourdata & en, es, hi, id \\
    \kbdata & 56 languages \\
    \end{tabular}
    \caption{The languages of each testing corpora in our experiments. The first three sets primarily include codemixed texts while the last one (\kbdata) is monolingual.}
	\label{tb:languages}
\end{table}

\paragraph{Evaluation datasets.}
We evaluate on four datasets, three codemixed and one monolingual. For a fair comparison, we report accuracies on subsets of these test sets that include languages supported by all tested models. Examples with Hindi words written in Latin script are also removed because the benchmark systems we compare to do not support it.

\begin{itemize}\itemsep0pt
\item \tb{\twitterdata}: Codemixed data from the EMNLP 2016 shared task \cite{Molina:2016}. 
\item \tb{\webdata}: Codemixed data crawled from multilingual web pages \citep{King:2013}, using a subset of six languages.
\item \textbf{\ourdata}: The test set of our token-level codemixed data (en-es, en-hi, and en-id). 
\item \tb{\kbdata:} Monolingual test set of \newcite{Kocmi:2017}, using a subset of 56 languages. 

\end{itemize}

\noindent Table \ref{tb:languages} summarizes the final language setting of each test set used in our experiments.

%% file: model.tex
\begin{figure}[t]
\centering
\includegraphics[width=0.45\textwidth]{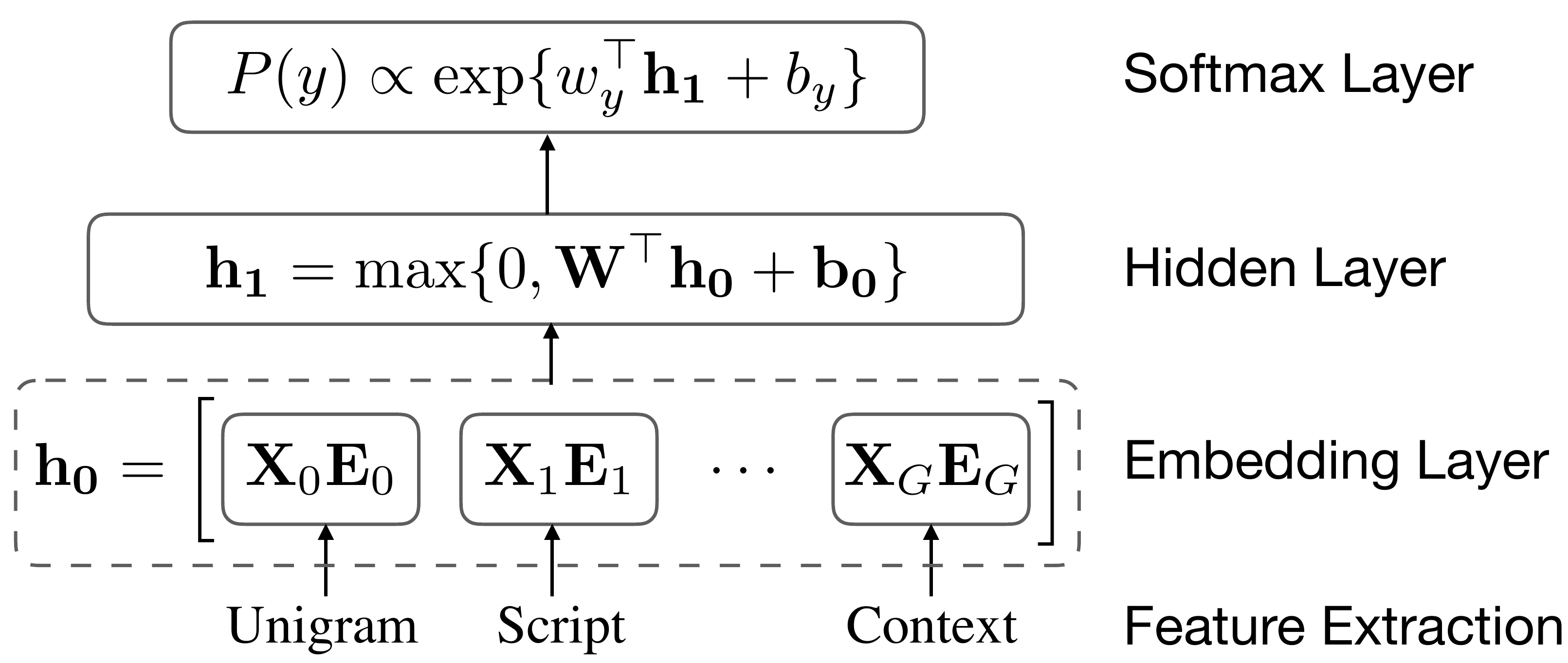}
\caption{Basic feed-forward network unit for scoring each token in the input and predicting possible languages. Multiple features are embedded, concatenated, and fed into a hidden layer with ReLU activation. 
}\label{fig:feedforward}
\end{figure}

\section{Identifying Language Spans in Codemixed Text}
\label{sec:model}

\mname\ uses two stages to assign language codes to every token in a sentence. First, it predicts a distribution over labels for each token independently with a feed-forward network that uses character and token features from a local context window. Then, it finds the best assignment of token labels for an entire sentence using greedy search, subject to a set of global constraints. Compared to sequence models like CRFs or RNNs, this two-stage strategy has several major advantages for fine-grained language ID: (1) it does not require annotated codemixed text over hundreds of languages and their mixed pairings, (2) learning independent classifiers followed by greedy decoding is significantly faster than structured training (especially considering the large label set inherent in language ID), and (3) it is far easier to implement.

\subsection{Token Model}
\label{sec:feedforward}

Simple feed-forward networks have achieved near state-of-the-art performance in a wide range of NLP tasks \cite{Botha:2017,Weiss:2015}. \mname\ follows this strategy, with embedding, hidden, and softmax layers as shown in Figure \ref{fig:feedforward}. The inputs to the network are grouped feature matrices, e.g. character, script and lexicon features. Each group $g$'s features are represented by a sparse matrix $\mathbf{X}_g\in\mathbb{R}^{F_g\times V_g}$, where $F_g$ is the number of feature templates and $V_g$ is the vocabulary size of the feature group. The network maps sparse features to dense embedding vectors and concatenates them to form the embedding layer:
\vspace{-0.5em}
\begin{equation}
\mathbf{h}_0 = vec[\mathbf{X}_g\mathbf{E}_g| \forall g]
\vspace{-0.5em}
\end{equation}

\noindent
where $\mathbf{E}_g\in\mathbb{R}^{V_g \times D_g}$ is a learned embedding matrix per group. The final size of the embedding layer $|\mathbf{h}_0| = \sum_g F_gD_g$ is the sum of all embedded feature sizes. \mname\ uses both discrete and continuous features. We use a single hidden layer with size 256 and apply a rectified linear unit (ReLU) over hidden layer outputs. A final softmax layer outputs probabilities for each language. The network is trained per-token with cross-entropy loss.


\begin{table}[t]
    \centering
    \begin{tabular}{@{~}l@{~}@{~}ccc@{~}}
    Features & Window & $D$ & $V$ \\
    \midrule
    Character $n$-gram & +/- 1 & 16 & 1000-5000\\
    Script & 0 & 8 & 27 \\
    Lexicon & +/- 1 & 16 & 100 \\
    \end{tabular}
    \caption{Feature spaces of \mname. The window column indicates that \mname\ uses character $n$-gram and lexicon features extracted from the previous and following tokens as well as the current one.}
	\label{tb:features}
\end{table}

\label{sec:features}
We explain the extraction process of each feature type below. Table \ref{tb:features} summarizes the three types of features and their sizes used in \mname. Character and lexicon features are extracted for the previous and following tokens as well as the current token to provide additional context.

\paragraph{Character $n$-gram features}

We apply character $n$-gram features with $n=[1, 4]$. RNNs or CNNs would provide more flexible character feature representations, but our initial experiments did not show significant gains over simpler $n$-gram features. We use a distinct feature group for each $n$. The model averages the embeddings according to the fractions of each $n$-gram string in the input token. For example, if the token is {\it banana}, then one of the extracted trigrams is \textit{ana} and the corresponding fraction is 2/6. Note that there are six trigrams in total due to an additional boundary symbol at both ends of the token.

Following \newcite{Botha:2017}, we use feature hashing to control the size $V$ and avoid storing a big string-to-id map in memory during runtime. The feature id of an $n$-gram string $x$ is given by $\mathcal{H}(x)\!\!\!\!\mod V_g$ \cite{Ganchev:2008}, where $\mathcal{H}$ is a well-behaved hash function. We set $V=1000,1000,5000,5000$ for $n=1,2,3,4$ respectively; these values yield good performance and are far smaller than the number of $n$-gram types.

\paragraph{Script features} Some text scripts are strongly correlated with specific languages. For example, Hiragana is only used in Japanese and Hangul is only used in Korean. Each character is assigned one of the 27 types of scripts based on its unicode value. The final feature vector contains the normalized counts of all character scripts observed in the input token.

\paragraph{Lexicon features} This feature group is backed by a large lexicon table which holds a language distribution for each token observed in the monolingual training data. For example, the word {\it mango} occurs 48\% of the time in English documents and 18\% in Spanish ones. The table contains about 6.2 million entries. We also construct an additional prefix table of language distributions for $6$-gram character prefixes. If the input token matches an entry in the lexicon table (or failing that, the prefix table), our model extracts the following three groups of features.


\begin{itemize}[topsep=0pt]\itemsep0pt
\item {\it Language distribution}. The language distribution itself is included as the feature vector.
\item {\it Active languages}. As above, but feature values are set to 1 for all non-zero probabilities. For example, the word {\it mango} has feature value 1 on both English and Spanish.
\item {\it Singletons}. If the token is associated with only one language, return a one-hot vector whose only non-zero value is the position indicating that language.
\end{itemize}


\noindent
The size of all lexicon feature vectors is equal to the number of supported languages.



\begin{figure*}[t]
\centering
\includegraphics[width=0.90\textwidth]{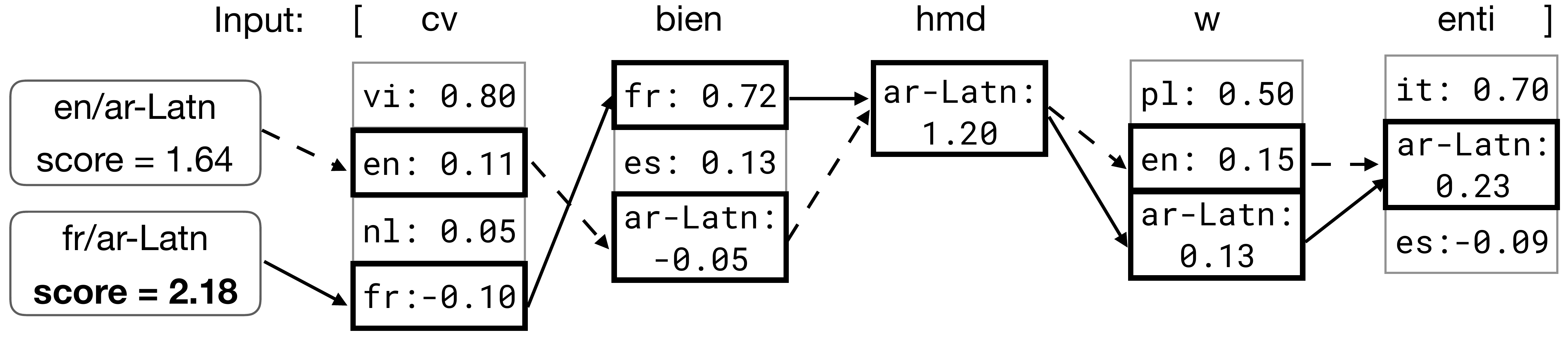}
\caption{Example of our decoding algorithm with global constraints for example \ref{ex:fr-ar} for two allowed language pairs, \mbox{en/ar-Latn} and \mbox{fr/ar-Latn}.
}\label{fig:decoding}
\end{figure*}

\begin{figure}[t]
\centering
\includegraphics[width=0.4\textwidth]{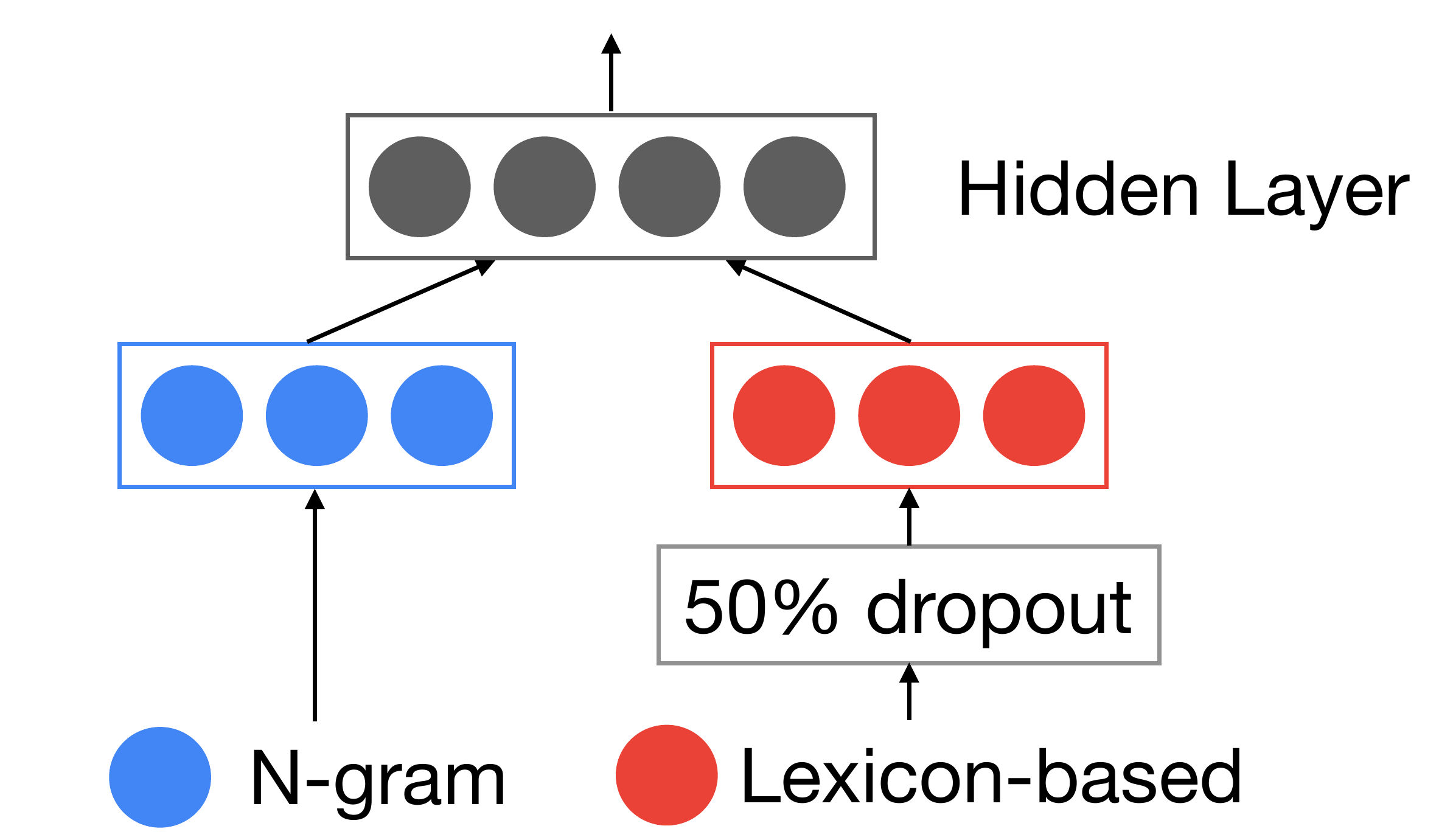}
\caption{Our selective feature dropout method. The model randomly sets the lexicon feature vectors to zero with 50\% probability while $n$-gram features are always used.
}\label{fig:dropout}
\end{figure}

\subsection{Selective Feature Dropout}
Preliminary experiments showed poor performance, especially on informal texts, when all three types of features are simply merged. Consider the following example outputs on misspelled word {\it Ennnglish}, for which no lexicon features fire.

\begin{center}
\begin{tabular}{l|l}
\multicolumn{2}{l}{Input: \textit{Ennnglish}} \\
\hline
{\small \textit{With Lexicon Features}{~~~}} & {\small \textit{W/o Lexicon Features}{~~~}}\\
{~~~~}\tabitem $p(sv) = 0.27$ & {~~~~}\tabitem $p(en) = 0.74$ \\
{~~~~}\tabitem $p(da) = 0.24$ & {~~~~}\tabitem $p(nl) = 0.10$ \\
{~~~~}\tabitem $p(nl) = 0.18$ & {~~~~}\tabitem $p(fy) = 0.06$ \\
{~~~~}\tabitem $\ldots$ & {~~~~}\tabitem $\ldots$ \\
\end{tabular}
\end{center}
Without dropout, the model with lexicon features does not make effective use of the token's character $n$-grams and makes a catastrophically wrong prediction. The core problem is that lexicon features are both prevalent and highly predictive for language ID; during training, this dampens the updating of weights of $n$-gram features and thus diminishes their overall utility.

To address this issue and make \mname\ more robust to noisy inputs, we selectively apply a grouped feature dropout strategy that stochastically down-weights lexicon features during training. Figure \ref{fig:dropout} illustrates the idea: for each input, after feature extraction, the vector of lexicon features is randomly set to zero. This way, the model must rely entirely on $n$-gram features for this particular input. Note that our feature dropout is different from standard dropout in at least two ways: (1) dropout happens to entire feature groups rather than on individual neurons, (2) we selectively apply dropout only on a subset of features. After tuning the dropout rate on development data (Figure \ref{fig:dropout_rate}) we choose a dropout rate of 50\%. Section \ref{sec:analysis} explains the tuning procedure.

\subsection{Decoding with Global Constraints} \label{sec:decoding}
Given a trained model, the goal of decoding is to find the sequence of per-token languages that maximizes the overall score. The simple, greedy strategy of picking the top prediction for each token over-predicts too many languages in a single sentence. For example, on average the greedy method predicts more than 1.7 languages per sentence on monolingual inputs. Because the token classier uses a window including only the previous, current, and next token, it has a quite limited view on the entire sequence. 

Motivated by this observation, we add the following \textbf{global constraint} in decoding: only monolingual outputs or codemixed outputs from a fixed set of language pairs are permitted. We choose a set of 100 language pairs, primarily including the combination of English and a non-English language. The full set is listed in the supplemental material.




\begin{table*}[t]
    \centering
    \begin{tabular}{lcccccccc}
	\toprule
    \textsc{Dataset} & \twitterdata & \webdata &  \multicolumn{4}{c} {\ourdata} & \multirow{2}{*}{{~~}Average} \\
    \cmidrule{1-7} 
    \textsc{Languages} &  {~~}es/en{~~} & 6 Langs & {~~}es/en{~~} & {~~}hi/en{~~} & hi/hi-Latn/en & {~~}id/en{~~} &\\
    \midrule
    \lanidenn & 71.3 & 52.1  & 65.7 & 79.6 & -- & 22.9  &  58.3\\
    \equilid  & 87.9 & 63.5 & 71.0 & 81.9 & -- & 64.9   & 73.9 \\
    \msmall & 88.8 & 91.0 & 89.9 & 98.2 & 85.0 & 86.7  &  90.9 \\
    \mfull &  \tb{92.4} & \tb{93.2} & \tb{91.8} & \tb{98.4} & \tb{87.4} & \tb{91.1} &  \tb{93.4}\\
    \bottomrule
    \end{tabular}
    \caption{\textbf{Codemixed Texts:} Token-level accuracy (\%) of different approaches on codemixed texts. ``\msmall'' corresponds to our small model without lexicon features and vocabulary tables. The \textit{hi/hi-Latn/en} column shows the accuracy on texts in English, Latin Hindi and Devanagari Hindi; the baseline models do not support identification of text in Hindi in Latin script. \textit{Average} shows averaged accuracy on all sets except \textit{hi/hi-Latn/en}. Boldface numbers indicate the best accuracy for each testing set.}
	\label{tb:codemixed_results}
\end{table*}

Finally, we introduce a straightforward variant of greedy decoding that finds the optimal language assignment in the presence of these global constraints. We independently find the best assignment under each allowed language combination (monolingual or language pair) and return the one with the highest score. 

Figure \ref{fig:decoding} shows paths for example \ref{ex:fr-ar} with two allowed language pairs: \mbox{en/ar-Latn} and \mbox{fr/ar-Latn}.\footnote{Scores are sorted. Some languages omitted for illustration purposes.} The two paths in dashed and solid lines indicate the best assignment for each language pair respectively. Because scoring is independent across tokens, each subtask is computed in $O(N)$ time. The total decoding time is $O(N|\mathcal{L}|)$ where $\mathcal{L}$ is the constraint set, and the global optimality of this algorithm is guaranteed because the assignment found in each subtask is optimal.



%% file: results.tex
\section{Experiments} \label{sec:experiments}
\subsection{Training Setup}
We train \mname\ on the concatenation of three datasets: (a) \ourdata's training portion, (b) synthetic codemixed data and (c) a monolingual corpus that covers 100 languages. Every token in the training set spawns a training instance. Our training set consists of 38M tokens in total, which is on the same magnitude as the sizes of training data reported in previous work \cite{jurgens2017incorporating,joulin2016fasttext}.

We use mini-batched averaged stochastic gradient descent (ASGD) \cite{Bottou:2010} with momentum \cite{Hinton:2012} and exponentially decaying learning rates to learn the parameters of the network. We fix the mini-batch size to 256 and the momentum rate to 0.9. We tune the initial learning rate and the decay step using development data.



\subsection{Main Results}
\paragraph{Codemixed Texts} Table \ref{tb:codemixed_results} lists our main results on the codemixed datasets. We primarily compare our approach against two benchmark systems: \equilid\ \cite{jurgens2017incorporating} and \lanidenn\  \cite{Kocmi:2017}. Both achieved state-of-the-art performance on several monolingual and codemixed language ID datasets. \lanidenn\ makes a prediction for every character, so we convert its outputs to per-token predictions by a voting method over characters in each word. For both benchmarks, we use the public pre-trained model provided by the authors. The \equilid\ model uses 53M parameters, \lanidenn\ uses 3M, and \mname\ only uses 0.28M parameters.\footnote{We explain how we compute the number of parameters of our model in the supplemental material.}  


Across all datasets, \mname\ consistently outperforms both benchmark systems by a large margin. On average, our full model (\mfull) is 19.5\%  more accurate than \equilid\ (93.4\% vs. 73.9\%); the gain is even larger compared to \lanidenn. Note that none of the models are trained on the \twitterdata\ or the \webdata\ dataset, so these two datasets provide an evaluation on the out-domain performance of each approach. In this setting \mfull\ also yields significant improvement in accuracy, e.g. a 4.5\% (absolute) gain over \equilid\ on the \twitterdata\ dataset.

\paragraph{An Even Smaller Model} We further compare between \mfull\ and a variant we call \msmall, which has no access to lexicon resources or lexicon features. This smaller variant has only 237k parameters and reduces the memory footprint from 30M to 0.9M during runtime, while the (average) loss on accuracy is only 2.5\%. This comparison demonstrates that our approach is also an excellent fit for resource-constrained environments, such as on mobile phones.

\begin{table}[t]
    \centering
    \begin{tabular}{lcr}
	\toprule
    \textsc{Model} & Sent Acc. & Char/Sec \\
    \midrule
    \multicolumn{2}{l}{\textsc{Codemixing Models}} & \\
    {\quad}\lanidenn  & 94.6 & 0.17k \\
    {\quad}\equilid  & 95.1 & 0.25k \\
    {\quad}\msmall & 94.6 & 265.5k  \\
    {\quad}\mfull & \tb{96.6} & 206.1k  \\
    \midrule
    \multicolumn{2}{l}{\textsc{Monolingual Models}} & \\
    {\quad}Langid.py  & 92.8 & 183.8k \\
    {\quad}fastText-small  & 92.5 & 2,671.1k \\
    {\quad}fastText-full & 94.4 & 2,428.3k \\
    {\quad}CLD2  & 95.5 & 4,355.0k \\
    \bottomrule
    \end{tabular}
    \caption{\textbf{Monolingual Texts:} Sentence-level accuracy (\%) on \kbdata. Monolingual models make per-sentence predictions only.}
	\label{tb:monolingual_results}
\end{table}


\paragraph{Monolingual Texts} In addition to \equilid\ and \lanidenn, we further compare \mname\ against Langid.py \cite{Lui:2012}, CLD2\footnote{\url{https://github.com/CLD2Owners/cld2}} and fastText \cite{joulin2016fasttext,joulin2016bag}---all are popular off-the-shelf tools for monolingual language ID. Sentence-level predictions for \equilid\ and \lanidenn\ models are obtained by simple voting. 
Table \ref{tb:monolingual_results} summarizes sentence-level accuracy
of different approaches on the \kbdata\ test set.
\mfull\ achieves the best sentence-level accuracy over all monolingual and codemixing benchmark systems. The resource-constrained \msmall\ also performs strongly, obtaining 94.6\% accuracy on this test set.

\begin{figure}[t]
\centering
\includegraphics[width=0.45\textwidth]{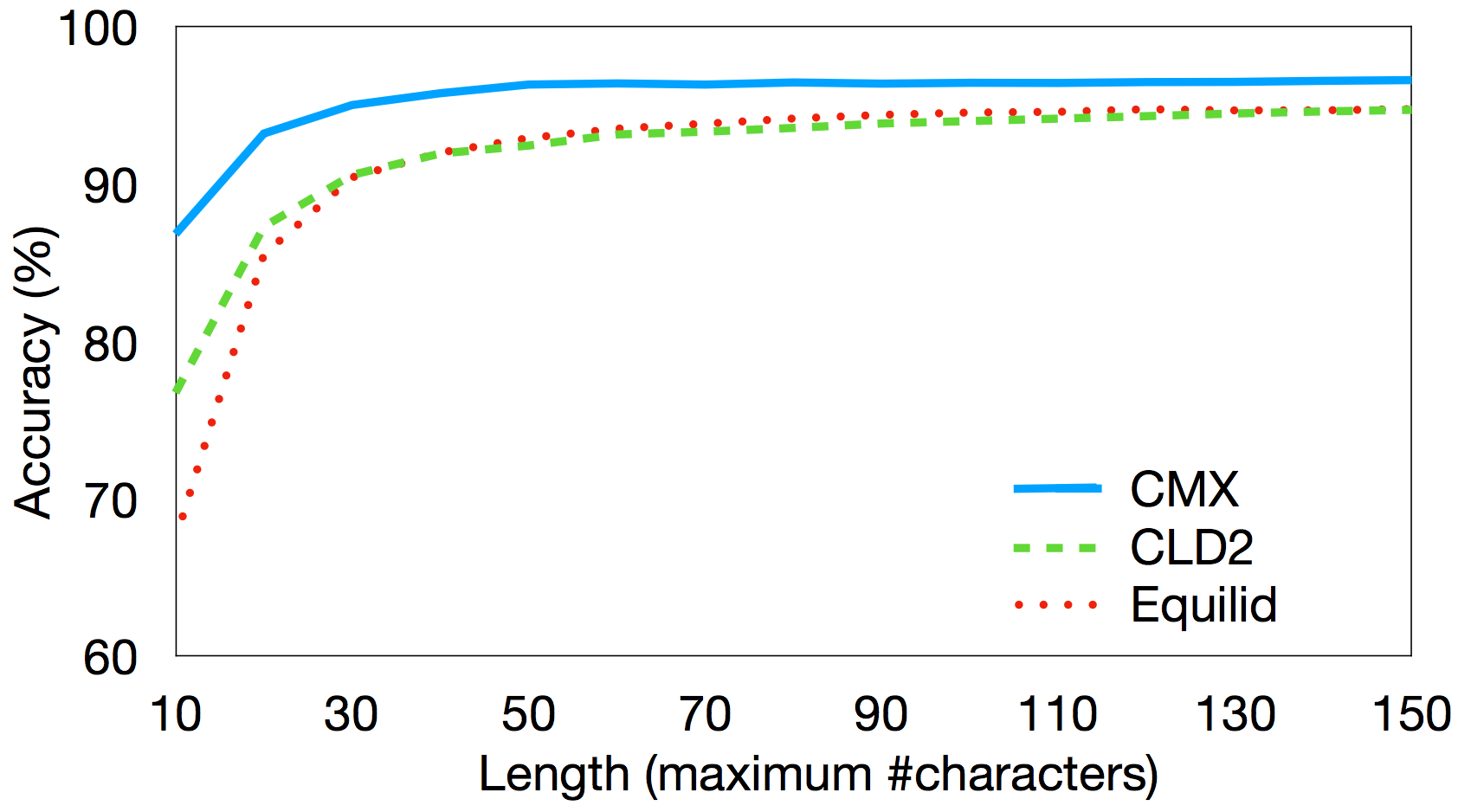}
\caption{Sentence-level accuracy ($y$-axis) on \kbdata\ as a function of the maximum number of non-whitespace characters in a sentence ($x$-axis). For example, the point at $x=50$ denotes the accuracy on all the sentences with $\le 50$ characters.
}\label{fig:length_accum}
\end{figure}

Our approach also maintains high performance on very short texts, which is especially important for many language identification contexts such as user-generated content. This is demonstrated in Figure \ref{fig:length_accum}, which plots the cumulative accuracy curve on \kbdata\ over sentence length (as measured by the number of non-whitespace characters). For example, points at $x{=}50$ show the averaged accuracies over sentences with no more than 50 characters. We compare \mfull\ against the best performing monolingual and codemixing benchmark systems. The relative gain is more prominent on shorter sentences than on longer ones. For example, the improvement is 4.6\% on short sentences ($\le$30 characters), while the gain on segments $\le$150 characters is 1.9\%. Similar patterns are seen with respect to other systems. 


\paragraph{Inference Speed} Table \ref{tb:monolingual_results} also shows the inference speed of each method in characters per second, tested on a machine with a 3.5GHz Intel Xeon processor and 32GB RAM. \mname\ (written in C++) is far faster than other fine-grained systems, e.g. it has an 800x speed-up over \equilid. It is not surprising that monolingual models are faster than \mname, which makes a prediction for every token rather than once for the entire sequence. Of course, monolingual models do not support language ID on codemixed inputs, and furthermore \mname\ performs the best even on monolingual texts.

\subsection{Analysis}\label{sec:analysis}

\begin{figure}[t]
\centering
\includegraphics[width=0.45\textwidth]{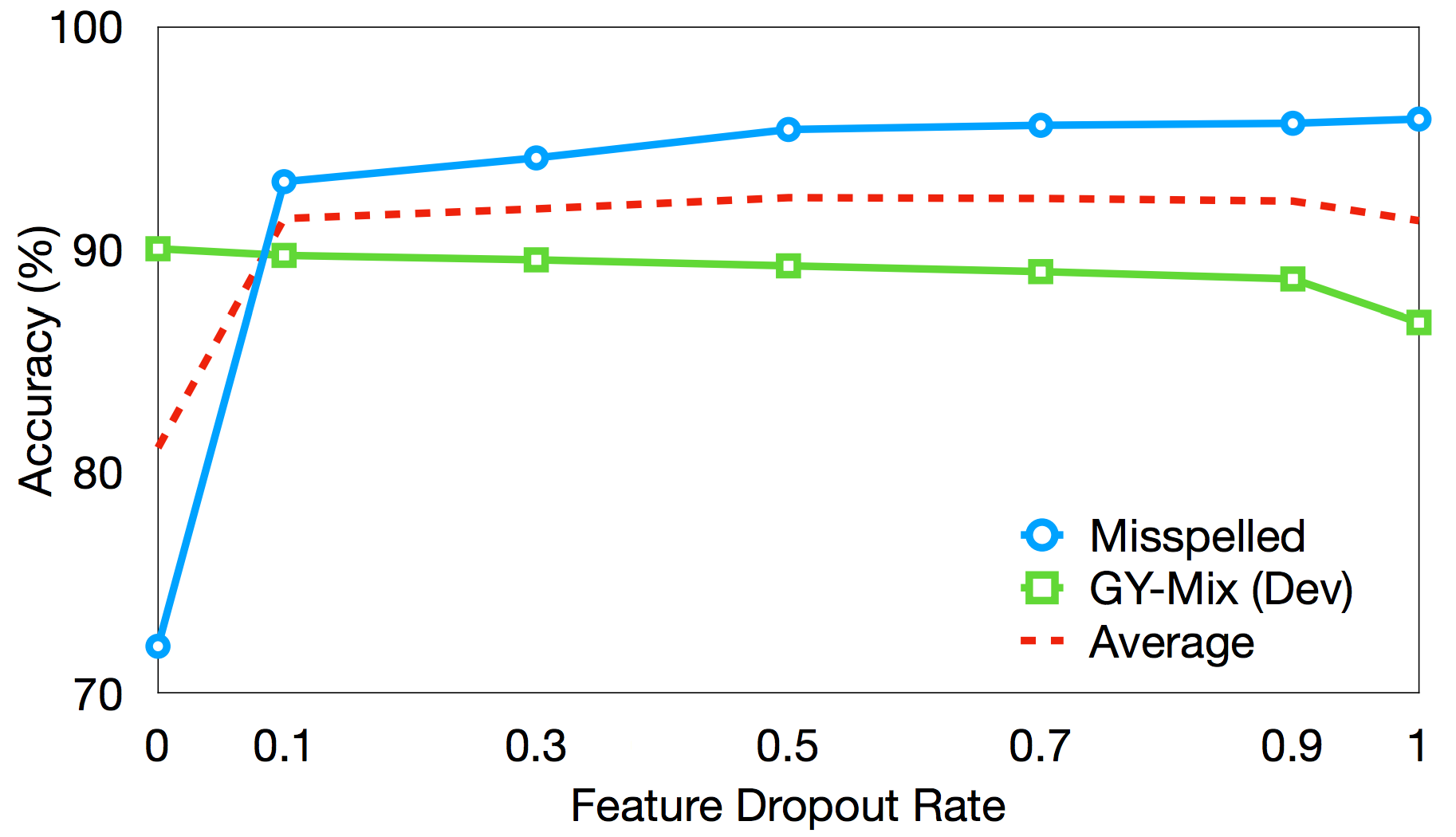}
\caption{Accuracy on development sets with various feature dropout rate values $p$.
}\label{fig:dropout_rate}
\end{figure}

\begin{table*}[t]
    \centering
    \begin{tabular}{lccccc}
	\toprule
    \textsc{Training Data} & \twitterdata & \webdata &\ourdata\ (Test) & \kbdata \\
    \midrule
    All Training Corpora & 92.4  & 93.2 & 93.6 & 95.1 \\
    w/o \ourdata\ (Train) & 88.5 & 92.9 & 89.3 & 95.0 \\
    w/o Synthetic & 92.1 & 88.8 & 92.5 & 95.1 \\
    \bottomrule
    \end{tabular}
    \caption{Token-level accuracy of our full model with different training sets, removing either \ourdata\ annotations or synthetic codemixed corpus at a time.}
	\label{tb:training_data}
\end{table*}

\begin{table}[t]
    \small
    \centering
    \begin{tabular}{@{}l@{~}@{~}c@{~}@{~}c@{~}@{~}c@{}}
	\toprule
    \textsc{Method} & \ourdata & \kbdata & \#Lang/Sent \\
    \midrule
    Independent & 87.8 & 91.9 & 1.78 \\
    Switching Penalty & 89.4 & 93.1 & 1.58   \\
    Bilingually Constrained & 93.6 & 95.1 & 1.27  \\
    \midrule
    Gold & - & - & 1.15 \\
    \bottomrule
    \end{tabular}
    \caption{Token-level accuracy of different decoding methods on \ourdata\ and \kbdata, as well as the averaged number of predicted languages in each sentence.}
	\label{tb:decoding}
\end{table}

\paragraph{Feature Dropout Rate} To analyze how the feature dropout rate impacts the model performance, we create a set of \textit{synthetically misspelled} tokens by random duplication or replacement of one or two characters. In addition, we ensure that every token has at least one language-unique character, so a model with character $n$-gram features should be able to easily identify the language of this token. Figure \ref{fig:dropout_rate} shows the tuning results for dropout values on misspelled tokens and the \ourdata\ development set. Without feature dropout ($p{=}0.0$), our model only gets 72.1\% on misspelled tokens, indicating that $n$-gram features are not properly trained. The proposed feature dropout method effectively addresses this issue, improving the accuracy to 95.3\% with $p \ge 0.5$. We choose $p=0.5$ (Figure \ref{fig:dropout}) because it gives the best trade-off on the two tuning sets. The curves in Figure \ref{fig:dropout_rate} also show that model performance is robust across a wide range of dropout rates between the two extremes, so the strategy is effective, but is not highly sensitive and does not require careful tuning. 

\vspace{-0.2em}
\paragraph{Impact of Decoding Algorithm}
Table \ref{tb:decoding} shows a comparison over different decoding strategies, including (a) \textit{independent} greedy prediction for each token, (b) adding a \textit{switching penalty} and decoding with Viterbi, (c) and our \textit{bilingually constrained} decoding. For the second method, we add a fixed transition matrix that gives a penalty score $\log{p}$ for every code switch in a sentence. We choose $p=0.5$, which gives the best overall results on the development set. Our approach outperforms \textit{switching penalty} by more than 2\% on both \ourdata\ and \kbdata. To analyze the reason behind this difference we show the average number of languages in each sentence in Table \ref{tb:decoding}. Both baseline approaches on average predict more than 1.5 languages per sentence while the oracle number based on gold labels is only 1.15. Our global bilingual constraints effectively address this over-prediction issue, reducing the average number of predicted languages to 1.27. We also measure the running time of all methods. The decoding speed of our method is 206k char/sec (Table \ref{tb:monolingual_results}), while the \textit{independent} method is 220k char/sec. Our decoding with global constraints thus only increases the running time by a factor of 1.07.



\paragraph{Codemixed Training Datasets}
Our training data consists of two codemixed corpora: manual annotations on real-world data (\ourdata) and a synthetic corpus. To analyze their contribution, we remove each corpus in turn from the training set and report the results in Table \ref{tb:training_data}. Adding the \ourdata\ training set mainly improves accuracy on \ourdata\ test and \twitterdata, while the gains from the synthetic data are greatest on \webdata. This shows that synthetic data helps \mname\ generalize to a broader range of languages since \ourdata\ has language overlap only with \twitterdata, not \webdata. 
The two examples below further demonstrate the benefit of synthetic examples:

{
\begin{center}
\begin{tabular}{l}
\textit{With Synthetic Data} \\
{~~} \tabitem [ Translate ]$_{\mathrm{en}}$ [ \tb{ma\c{c}\~{a}} ]$_{\mathrm{pt}}$ [ to English ]$_{\mathrm{en}}$  \\
{~~} \tabitem [ Translate ]$_{\mathrm{en}}$ [ \tb{Apfel} ]$_{\mathrm{de}}$ [ to English ]$_{\mathrm{en}}$    \\
\textit{Without Synthetic Data} \\
{~~} \tabitem [ Translate  \tb{ma\c{c}\~{a}}  to ]$_{\mathrm{pt}}$ [ English ]$_{\mathrm{en}}$  \\
{~~} \tabitem [ Translate \tb{Apfel} to English ]$_{\mathrm{en}}$    \\
\end{tabular}
\end{center}
}

Both examples are likely potential queries ``Translate \textit{apple} to English'' with \textit{apple} replaced by its translation in German(de) or Portuguese(pt). The underlying language pairings never appear in \ourdata. \mname\ with synthetic training data is able to correctly identify the single token inter-mixed in a sentence, while the model trained without synthetic data fails on both cases.


\paragraph{Contribution of Features} \mname\ has three types of features: character $n$-gram, script, and lexicon features. $n$-gram features play a crucial role as back-off from lexicon features. Consider informal Latin script inputs, like \textit{hellooooo}, for which no lexicon features fire. Foregoing $n$-gram features results in abysmal performance ($<$20\%) on this type of input because script features alone are inadequate. The main impact of script features is to avoid embarrassing mistakes on inputs that can be easily identified from their scripts. 
Finally, note that removing lexicon features corresponds to the \msmall\ model. On monolingual inputs (Table \ref{tb:monolingual_results}), the lexicon features in \mfull\ provide a 2.0\% absolute improvement in accuracy.

%% file: conclusion.tex
\section{Conclusions}

\mname\ is a fast and compact model for fine-grained language identification.  It outperforms related models on codemixed and monolingual texts, which we show on several datasets covering text in a variety of languages and gathered from diverse sources. Furthermore, it is particularly robust to the idiosyncrasies of short informal text.